\documentclass{article}
\usepackage{spconf,amsmath,graphicx}
\usepackage{amsmath,graphicx}

\usepackage{braket}
\usepackage{mathtools}
\usepackage{amsmath}
\usepackage{amssymb}
\usepackage{amsfonts}
\usepackage{lipsum}
\usepackage{lscape}
\usepackage{multirow}
\usepackage{here}
\usepackage{bm}
\usepackage{color}
\usepackage{cuted}
\usepackage{adjustbox}
\usepackage{cite}
\usepackage{tabularx}
\usepackage{enumitem}
\usepackage{booktabs}
\usepackage{sistyle}
\SIthousandsep{,}

\usepackage{setspace}

\usepackage{here}
\usepackage{subfigure}
\usepackage{tabularray}
\usepackage{amsthm}

\usepackage{array}

\usepackage{nicematrix}

\usepackage[skip=0pt]{caption}

\usepackage[ruled,vlined]{algorithm2e}

\SetKwInput{KwInput}{Input}                %
\SetKwInput{KwOutput}{Output}

\newcommand{\mb}[1]{\mathbf{#1}}
\newcommand{\tr}{\mathop{\mathrm{tr}}}

\newcommand{\argmax}{\mathop{\rm arg~max}\limits}
\newcommand{\argmin}{\mathop{\rm arg~min}\limits}

\title{Optimizing $\MakeLowercase{k}$ in $\MakeLowercase{k}$NN graphs with graph learning perspective}
\name{Asuka Tamaru$^{1\ast}$, Junya Hara$^{2\ast}$, Hiroshi Higashi$^2$, Yuichi Tanaka$^{2}$, and Antonio Ortega$^3$ \thanks{$\ast$: Equal contribution. This work was supported in part by JSPS KAKENHI under Grant 23H01415, 22H05163, and 22K12500, JST AdCORP under Grant JPMJKB2307, and by the US NSF under grant CCF-2009032.}}
\address{$^1$Tokyo University of Agriculture and Technology, Tokyo, Japan\\
$^2$Osaka University, Osaka, Japan\\
$^3$University of Southern California, Los Angeles, CA}
\begin{document}
\ninept
\maketitle
\begin{abstract}
In this paper, we propose a method, based on graph signal processing, to optimize the choice of $k$ in $k$-nearest neighbor graphs ($k$NNGs).
$k$NN is one of the most popular approaches and is widely used in machine learning and signal processing.
The parameter $k$ represents the number of neighbors that are connected to the target node; however,
its appropriate selection is still a challenging problem.
Therefore, most $k$NNGs use ad hoc selection methods for $k$.
In the proposed method, we assume that a different $k$ can be chosen for each node. 
We formulate a discrete optimization problem to seek the best $k$ with a constraint on the sum of distances of the connected nodes.
The optimal $k$ values are efficiently obtained without solving a complex optimization.
Furthermore, we reveal that the proposed method is closely related to existing graph learning methods.
In experiments on real datasets,
we demonstrate that the $k$NNGs obtained with our method are sparse and can determine an appropriate variable number of edges per node.
We validate the effectiveness of the proposed method for point cloud denoising, comparing our denoising performance with achievable graph construction methods that can be scaled to typical point cloud sizes (e.g., thousands of nodes). 
\end{abstract}

\section{Introduction}
\label{sec:intro}
Graphs are a powerful mathematical tool in machine learning and signal processing and can explicitly express complex inter-node relationships \cite{ortega_graph_2018,shuman_emerging_2013,tanaka_sampling_2020,cheung_graph_2018}, in scenarios where data are distributed irregularly in space or can be assumed to have some underlying graph structure, e.g., social, sensor, power, and brain networks \cite{mateos_connecting_2019,dong_graph_2020,jablonski_graph_2017}.

Graphs are sometimes known \textit{a priori}, e.g., transportation networks or power grids.
However, in many applications, such as sensor networks or point clouds, graphs are not given and need to be learned or estimated from given data or prior knowledge so that some requirements (e.g., average degree or distance) are satisfied, by solving a graph learning/estimation problem \cite{dong_learning_2019,mateos_connecting_2019,dong_graph_2020, kalofolias_large_2019,muja_scalable_2014}.
Graphs having sparse edge distributions are usually preferred since they lead to more explainable structures and require less computational cost for downstream applications.
Representative graph learning/estimation methods focus on learning a sparse graph from data without losing important topological characteristics\cite{egilmez_graph_2019,shekkizhar_graph_2019,zeng_sparse_2022,friedman_sparse_2008,dong_learning_2016}.

\begin{figure}[t!]
\centering
 \subfigure[][Fixed $k$NN]
  {\centering\includegraphics[width=0.35\linewidth]{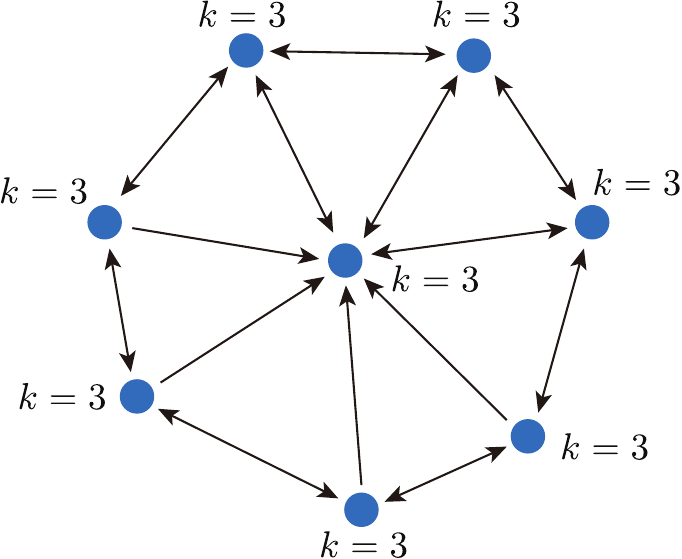}}\hspace{1em}
 \subfigure[][Variable $k$NN]
  {\centering\includegraphics[width=0.35\linewidth]{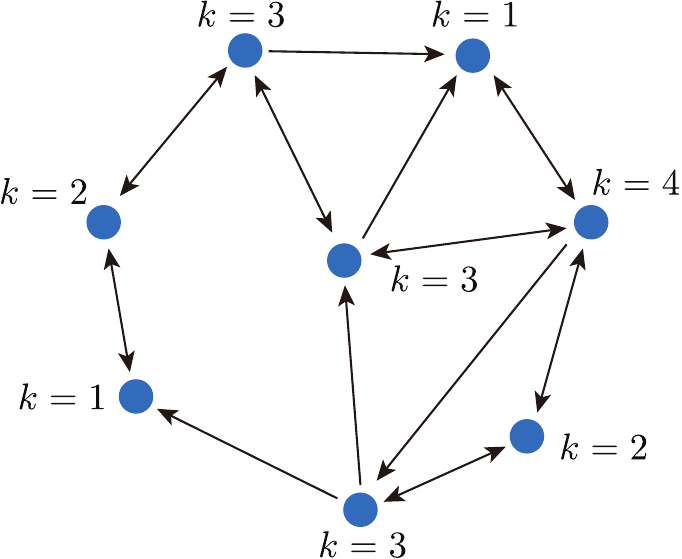}}
\caption{Comparison of $k$NNGs.}
\vspace{-0.5cm}
\label{fig:comparison_fixed_variable_knng}
\end{figure}

The $k$-nearest neighbor graph ($k$NNG) is one of the most popular graph construction methods. 
Each node in a $k$NNG is connected to its nearest $k$ nodes, where some pre-selected metric measures the distance.
Since $k$ is determined prior to the graph construction, $k$NNGs are guaranteed to have sparse connectivity as long as $k$ is small. 
However, if we use the fixed $k$, edge sparseness is automatically fixed for all nodes, which may reduce explainability and impact performance in some applications.
For example, if $\ell > k$ nodes form a dense cluster, $(\ell - k)$ nodes cannot be connected to the target node.
Similarly, if the underlying graph is a tree, a large $k$ does not yield expected results in applications.
This has led to approaches where $k$ can differ for each node, i.e., $k$NNGs with varying $k$s  (v$k$NNGs) \cite{zhang_efficient_2018}.
The difference between $k$NNG and v$k$NNG is depicted in Fig.~\ref{fig:comparison_fixed_variable_knng}.
The main challenge in v$k$NNGs design has been a lack of 
criteria for determining $k$, which has led most existing methods to select $k$ in an \textit{ad hoc} manner \cite{qi_3d_2017,mills_segmentation_2007}.

Graph learning methods developed in a graph signal processing (GSP) context \cite{dong_learning_2019,mateos_connecting_2019,dong_graph_2020,yamada_temporal_2021, yamada_time-varying_2019} aim to learn a graph from a set of signals, with the goal of selecting edges so that the graph operator (e.g., the graph Laplacian) is the precision matrix (the inverse covariance) of a Gauss Markov random field with covariance that approximates the empirical covariance of the observed signals. 
Typically, this graph learning problem is formulated as a sparse precision matrix estimation with some regularization terms on the \textit{edge weights} \cite{egilmez_graph_2017,zeng_sparse_2022,friedman_sparse_2008,yamada_temporal_2021, yamada_time-varying_2019}.
These approaches are also referred to as \textit{sparse covariance matrix estimation} (SCME) \cite{friedman_sparse_2008}.
However, this optimization often produces too many edges and typically results in high computation complexity.
This is because it is inherently given by the continuous optimization with respect to the fully connected adjacency matrix, i.e., $N\times N$ matrix, even if a sparsity constraint is imposed on the adjacency matrix.
Therefore, existing graph learning methods could not be scaled for large dimensions.

This paper proposes a construction method for v$k$NNGs using a graph learning perspective.
We formulate a v$k$NNG construction problem as a discrete optimization that seeks the best $k$ nodes to connect with a constraint on the sum of distances.
This problem can be solved efficiently via node-wise optimization.
Furthermore, we reveal the relationship between the proposed v$k$NNG construction method and graph learning by showing that the proposed method can be interpreted as an approximation to SCME-based graph learning.
Our proposed approach can be viewed as a fast heuristic algorithm for graph learning that initializes the graph with a fixed $k$NNG, inheriting a computation efficiency comparable to that of the standard $k$NNG construction.
We also derive the computational complexity of the proposed method and show that our algorithm has much lower computation costs than existing SCME and v$k$NNG constructions.
In point cloud denoising, 
the proposed method shows superior denoising performance to existing methods for all objects.

\textit{Notation:} Bold lower and upper case letters represent a vector and a matrix, respectively.
We denote an $\ell_p$ norm by $\|\cdot\|_p$.
$[\mb{A}]_{i,j}$ and $(\mb{A})_{i}$ denote the $(i,j)$th element of $\mb{A}$ and the $i$th column of $\mb{A}$, respectively.
The all-one vector is represented as $\bm{1}$. 
We denote the element-wise product by $\circ$.

\section{Brief Review of GSP and Graph Construction/Estimation Methods}\label{sec:related_work}

In this section, we first we describe the GSP basics used throughout this paper and then  
review the conventional $k$NNG algorithm and graph construction methods in GSP.

\subsection{GSP Basics}\label{subsec:gsp_basic}

We consider a graph $\mathcal{G}=(\mathcal{V,E})$,
where $\mathcal{V}$ and $\mathcal{E}$ represent sets of nodes and edges, respectively. The number of nodes is $N = |\mathcal{V}|$ unless otherwise specified. A weighted adjacency matrix of $\mathcal{G}$ is denoted by $\mb{W}$ where its $(m,n)$-element $w_{m,n}\geq 0$ is the edge weight between the $m$th and $n$th nodes; $w_{m,n}=0$ for unconnected nodes. We also use $\mb{A}\in\{0,1\}^{N\times N}$ as the unweighted adjacency matrix of $\mb{G}$, which is the binary counterpart of $\mb{W}$.
The degree matrix $\mathbf{D}$ is defined as $\mathbf{D}=\text{diag}\,(d_{0},d_{1},\ldots,d_{N-1})$, where $d_{m}=\sum_nw_{m,n}$ is the $m$th diagonal element.
We use the graph Laplacian $ \mathbf{L}\coloneqq \mathbf{D}-\mb{W}$ as a graph operator. 
A graph signal $\bm{x} \in \mathbb{R}^N$ is defined as a function $x:\mathcal{V}\rightarrow \mathbb{R}$, i.e., $x[n]$ can be regarded as the signal value on the node $n$. We assume that a set of observed graph signals, represented by $\mb{X}=[\bm{x}_1, \dots,\bm{x}_D]\in\mathbb{R}^{N\times D}$, can be used for learning the graph. The empirical covariance matrix $\mb{\Sigma} = \mb{X} \mb{X}^\top$.

\subsection{$k$NNG Construction}

We assume that the parameter $k \in \mathbb{Z}^{+}$ and the observed graph signals $\mb{X}$ are given.
Formally, the standard (fixed) $k$NNG can be obtained by solving the following problem. 
\begin{equation}
\textstyle
\mb{A}_k=\argmax_{\mb{A}\in\{0,1\}^{N\times N}}\|\mb{A}\circ\bm{\mathcal{D}}\|_1\quad \text{s.t. }\sum_{j=1}^N[\mb{A}]_{i,j}=k,\label{eq:knn_graph}
\end{equation}
where $\bm{\mathcal{D}}$ is an arbitrary similarity matrix. We assume that $[\bm{\mathcal{D}}]_{i,j}$ is calculated with a positive definite kernel like cosine similarity and Gaussian radial basis function.

Similar to the fixed $k$ $k$NNG, a v$k$NNG is obtained by solving \eqref{eq:knn_graph} with respect to each column of $\mb{A}$. 
Let $\bm{a}_i \in \mathbb{R}^N$ be the $i$th column of $\mb{A}$.
Its cost function can be formulated as
\begin{align}
\bm{a}_{\text{variable}, k_i}=\argmax_{\bm{a}_i\in\{0,1\}^{N}}\|\bm{a}_i\circ (\bm{\mathcal{D}})_i\|_1\quad\text{s.t. }\bm{1}^\top\bm{a}_i=k_i,\label{eq:vknn_graph}
\end{align}
where $k_i$ is the number of connected nodes to the node $i$. Note that $\bm{a}_{\text{variable}, k_i}$ is identical to the $i$th column of $\mb{A}_k$, i.e., $\bm{a}_{\text{variable}, k_i}=(\mb{A}_k)_i$.
In \eqref{eq:vknn_graph}, we use the unweighted adjacency matrix $\mb{A}_k$.  
The weighted adjacency matrix of the fixed/variable $k$NNG can also be given by $\mb{W}_k=\mb{A}_k\circ\bm{\mathcal{D}}$.

These $k$NNGs have some advantages over the other methods, e.g., b-matching  \cite{huang_loopy_2007} and locally linear embedding 
\cite{roweis_nonlinear_2000}, given their easy implementation and low computational complexity.
Therefore, numerous signal processing and machine learning applications use $k$NNGs \cite{garcia_few-shot_2018,qi_3d_2017}.
On the other hand, determining the appropriate $k$ has been a challenging problem.

\subsection{Graph Estimation/Leaning}\label{subsec:gl}

Graph learning is based on latent models imposed on the graph \cite{dong_learning_2019,mateos_connecting_2019,dong_graph_2020}, such as the Gaussian Markov random field (GMRF)  model, whose precision matrix is given by $\mb{L}$ \cite{egilmez_graph_2017}. The  maximum likelihood estimator (MLE) of $\mb{L}$ can be obtained by solving \cite{friedman_sparse_2008}:
\begin{equation}
\begin{aligned}
\min_{\mb{L}} &\tr(\widetilde{\mb{L}}\mb{\Sigma})-\log\det(\widetilde{\mb{L}})+\mu\|\widetilde{\mb{L}}\|_1\\
\text{s.t. }&\widetilde{\mb{L}}=\mb{L}+\nu\mb{I},\mb{L}\bm{1}=\bm{0},[\mb{L}]_{i,j}\leq0\quad i\neq j,
\end{aligned}\label{eq:glasso}
\end{equation}
where $\nu>0$ is some constant and $\mu>0$ is a  parameter. The first and second terms are the fidelity for the MLE model, i.e., the Bregman divergence \cite{dhillon_matrix_2008}. The third term is an $\ell_1$-regularization, which encourages the sparsity for entries in $\mb{L}$. The constraints in \eqref{eq:glasso} ensure that the solution is a graph Laplacian. 
Since \eqref{eq:glasso} can be expressed as a semidefinite programming (SDP) problem, it can be solved by an iterative optimization.
However, its computational cost is generally much higher than the $k$NNG construction.

Non-negative kernel regression (NNK) is another graph estimation method \cite{shekkizhar_graph_2019}.
It aims to remove unnecessary edges from a $k$NNG and jointly optimize edge weights such that features on nodes are best approximated by the linear combination of the features on their neighbors.
The problem is solved by quadratic programming in a node-by-node manner.
Therefore, this can also be considered as one of the possible formulations of v$k$NNGs.

\section{V$\MakeLowercase{k}$NNG estimation}

In this section, we formulate the problem of determining $k$ in  v$k$NNGs and  
investigate the relationship between the graph learning approach introduced in Section~\ref{subsec:gl} and the proposed method. 
We also study the parameter selection in our method and compare its computational complexity to that of existing methods.

\subsection{Problem Formulation}
We start by modifying the formulation for the v$k$NNG construction in \eqref{eq:vknn_graph}. 
In this paper, we use the Euclidean distance matrix $\mb{Z}=(\mb{I}\circ\mb{\Sigma})\bm{1}\bm{1}^{\top}+\bm{1}\bm{1}^\top(\mb{I}\circ\mb{\Sigma})-2\mb{\Sigma}$ instead of the similarity matrix $\bm{\mathcal{D}}$  (see Sec.~\ref{subsec:gsp_basic}), which helps to understand the relationship between our method and graph learning shown later. With standard similarity measures, such as cosine similarity,  similarity and distance between two nodes are inversely proportional to each other. Hence, we can rewrite \eqref{eq:vknn_graph} as a maximization:
\begin{equation}
\bm{a}_{\text{variable}, k_i}=\argmin_{\bm{a}_i\in\{0,1\}^{N}}\|\bm{a}_i\circ (\mb{Z})_i\|_1\quad\text{s.t. }\bm{1}^\top\bm{a}_i=k_i,\label{eq:vknn_graph_2}, 
\end{equation}
which is equivalent to \eqref{eq:vknn_graph}: We need to determine $k_i$ a priori.

To tackle the problem, we swap the cost function and the constraint:
\begin{equation}
\bm{a}_{\text{variable}, k_i}=\argmax_{\bm{a}_i\in\{0,1\}^{N}}\|\bm{a}_i\|_1\quad\text{s.t. }\|\bm{a}_i\circ (\mb{Z})_i\|_1\leq \beta_i,\label{eq:new_proposed}
\end{equation}
where $\beta_i$ is a parameter for the vertex $i$.
This formulation removes the constraint on the degree and, thus, enables us to determine the optimal $k_i$ automatically.
Note that \eqref{eq:new_proposed} is easily solved because we only need to sequentially pick the indices corresponding to the largest elements in $\left( \mb{Z} \right)_i$, while verifying that the constraint is satisfied.

\begin{figure}[t!]
\centering
 \subfigure[][Fixed $k$NNG]
  {\centering\includegraphics[width=0.4\linewidth]{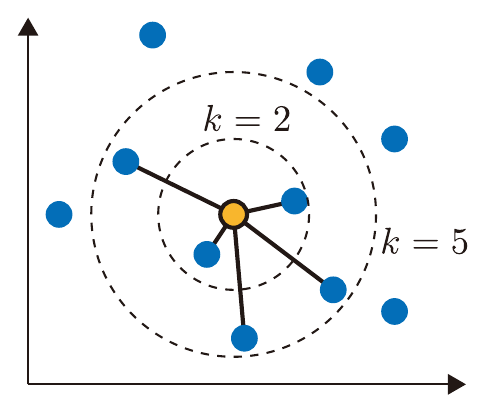}}\hspace{1em}
 \subfigure[][v$k$NNG]
  {\centering\includegraphics[width=0.4\linewidth]{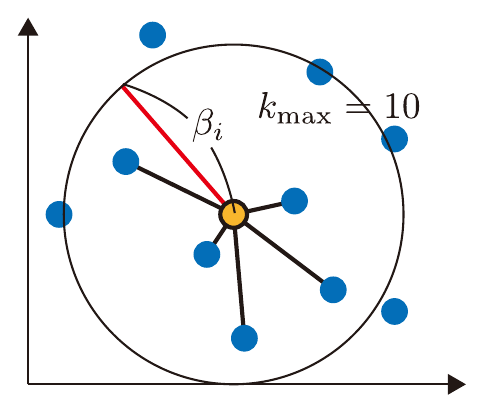}}
\caption{Geometric interpretation of the fixed and variable $k$NNGs in 2-D space. 
Note that $\beta_i$ in the v$k$NNG corresponds to the radius of the dotted bounding circle.}
\vspace{-0.5cm}
\label{fig:geo_interp}
\end{figure}

We visualize the geometric interpretation of the proposed method in Fig.~\ref{fig:geo_interp}.
Our approach can be viewed as an extension of $\epsilon$-nearest neighbor graphs ($\epsilon$-NNGs) with varying $k$s, which picks up the nodes included within the radius $\beta_i$.
Since the proposed method considers the sum of the distances $\beta_i$ instead of directly specifying the radius $\epsilon$, it can adapt to the distribution of nodes more flexibly than $\epsilon$-NNGs:
If we consider two similar nodes, where one is extremely close to the target node ($r_1<\epsilon$) and the other is relatively far from the target node ($r_2>\epsilon$), $\epsilon$-NNGs cannot connect the node with the radius $r_2$. On the other hand, the proposed method can connect both nodes if $r_1+r_2<\beta_i=2\epsilon$ is satisfied.

By comparing with \eqref{eq:knn_graph}, it is natural that the computational complexity of solving \eqref{eq:new_proposed} is compatible with that of the standard $k$NNG, which is detailed later.

The steps of our proposed method are shown in Algorithm \ref{algo:acc_proposed}.
Note that, with the spirit of $k$NNG, we set the minimum and maximum $k$s to avoid isolated nodes and/or too many edges.
Algorithm \ref{algo:acc_proposed} starts from the fixed $k_{\min}$NNG and \textit{adds} nodes which are sufficiently similar.
We can also starts from the fixed $k_{\max}$NNG and \textit{remove} dissimilar nodes.
They can be viewed as heuristics for initialization of graph learning with low calculation costs.

So far, there has been no clear relationship between $k$NNGs and graph learning.
In the following, we connect the graph learning presented in \eqref{eq:glasso} with the proposed v$k$NNG construction.

\subsection{Relationship to Graph Learning}
\label{sec:proposed}

Interestingly, we have a clear relationship between \eqref{eq:glasso} and \eqref{eq:new_proposed}.
We start by reformulating \eqref{eq:glasso} in terms of $\mb{W}$.
The first term in \eqref{eq:glasso} is rewritten as
\begin{equation}
\tr(\widetilde{\mb{L}}\mb{\Sigma})=\tr(\mb{L}\mb{\Sigma})+\nu\tr(\mb{\Sigma})=\|\mb{W}\circ \mb{Z}\|_1+\textit{const}.\label{eq:glasso_tr}
\end{equation}

We can also approximate the second term in \eqref{eq:glasso} with the Taylor series expansion and obtain the following form \cite{muir_theory_1890}:
\begin{equation}
\begin{split}
\log\det(\widetilde{\mb{L}})&=\tr(\mb{L})+\nu+T\left(\frac{1}{\nu}\right)
=\|\mb{W}\|_1+\nu+T\left(\frac{1}{\nu}\right),
\end{split}\label{eq:glasso_logdet}
\end{equation}
where $T$ is the remainder term.
Furthermore, the sparsity-promoting term in \eqref{eq:glasso} can be rewritten:
\begin{equation}
\begin{split}
\|\widetilde{\mb{L}}\|_1&
=2\|\mb{W}\|_1+\nu N.\label{eq:glasso_l1norm}
\end{split}
\end{equation}

\begin{algorithm}[t!]
  \KwInput{$\mb{Z}\in\mathbb{R}^{N\times N}$, $k_{\max}$, $k_{\min}$, $\beta_i$ ($i=1,2,\dots,N$)}
  \For{$i=1,2,\dots,N$}{
  Sort $\mb{Z}$ along columns in ascending order and set to $\mb{Z}^\star$\\
  Initialize $\Delta_{i}=[\mb{Z}^\star]_{i,k_{\min}}$ and $k_i= k_{\min}$\\
  \While{$(\Delta_{i}<\beta_i)$ $\land$ $(k_i<k_{\max})$}{
  $\Delta_{i}=\Delta_{i}+[\mb{Z}^\star]_{i,k}$\\
  $k_i=k_i+1$
  }
  }
  \KwOutput{$\{k_i^*\}_{i=1,\cdots,N}$}
\caption{Proposed v$k$NNG construction}\label{algo:acc_proposed}
\end{algorithm}

By ignoring the constant and perturbation terms in \eqref{eq:glasso_tr}, \eqref{eq:glasso_logdet}, and \eqref{eq:glasso_l1norm}, we can obtain an approximation of \eqref{eq:glasso} as follows:
\begin{equation}
\begin{aligned}
\min_{\mb{W}\in\mathbb{R}_{\ge 0}^{N\times N}} &\|\mb{W}\circ\mb{Z}\|_1-\alpha\|\mb{W}\|_1
\end{aligned}\label{eq:glasso_2}
\end{equation}
where $\alpha=1-2\mu$. Note that $\alpha$ is required to be positive, i.e., $\mu<\frac{1}{2}$.
Since \eqref{eq:glasso_2} is a difference-of-convex (DC) program, solving it directly is challenging.

\begin{table}[!t]
\centering
\caption{Comparison of computation complexities where GL refers to graph learning. 
$\omega$ is some constant factor for SDP.}\label{tab:complexity}
\begin{tabular}{c|c|c}
\hline
Method & Preparation  & Optimization\\\hline\hline
$k$NNG & $O(N^2 Dk_{\max})$ & - \\
GL (cf. \eqref{eq:glasso})& - & $O(N^5+N^{3\omega}+N^{2\omega})$\\
NNK\cite{shekkizhar_graph_2019} & $O(N^2 D k_{\max})$ & $O(Nk_{\max}^3)$\\
Proposed & $O(N^2Dk_{\min})$ & $O(N^2 D + N^2 + N k_{\max}))$\\\hline
\end{tabular}
\end{table}

Further, we divide the problem in \eqref{eq:glasso_2} into subproblems focusing on a target node.
We assume $\mathbf{W}$ is a binary matrix, i.e., $\mathbf{W} = \mathbf{A}$.
Since the $\ell_1$ norms in \eqref{eq:glasso_2} are separable across columns, we can transform \eqref{eq:glasso_2} into a v$k$NNG estimation as follows:
\begin{equation}
\bm{a}_{\text{variable}, k_i}=\argmin_{\bm{a}_i\in\{0,1\}^{N}}\|\bm{a}_i\circ (\mb{Z})_i\|_1-\alpha_i\|\bm{a}_i\|_1,\label{eq:old_proposed}
\end{equation}
where $\alpha_i$ is the parameter for the vertex $i$.

By considering a properly-chosen $\alpha_i$ corresponding to $\beta_i$, \eqref{eq:old_proposed} is nothing but \eqref{eq:new_proposed} (see \cite[Theorem 27.4]{rockafellar_convex_1970}). Therefore, we can see that \eqref{eq:old_proposed} and \eqref{eq:new_proposed} are essentially equivalent.

In summary, we can view the proposed v$k$NNG algorithm as a heuristic approach to solve a graph learning \eqref{eq:glasso} 
with much lower computational costs by focusing on the connection information.
As a result, our approach is closely related to graph learning both in algorithmic and theoretic senses.

\subsection{Parameter Selection}\label{subsec:param_select}

We now discuss the choice of the parameter $\beta_i$ in \eqref{eq:new_proposed}. To begin with, we consider the selection of $\alpha_i$ in \eqref{eq:old_proposed}. By rewriting the $\ell_1$ norm with trace, the equivalent form to \eqref{eq:old_proposed} can be derived as
\begin{equation}
k_i^* = \argmin_{k_i\in\mathcal{K}} \bm{a}_i^\top((\mathbf{Z})_i-\alpha_i\bm{1})\ \text{ for }i=1,\ldots,N.
\end{equation}
One of the reasonable choices of $\alpha_i$ is based on the geometric center of $(\mathbf{Z})_i$, which is defined as
$
(\mb{Z})_{\text{cent},i}\coloneqq (\mb{Z})_i-\frac{1}{N}\bm{1}(\mb{Z})_i^\top\bm{1}.
$
As a result, we recommend to set $\alpha_i= \frac{1}{N}(\mb{Z})_i^\top\bm{1} = \frac{1}{N}\sum_{j=0}^N[\mb{Z}]_{i,j}$: This could help to remove the bias.

Although the explicit relationship between $\alpha_i$ and $\beta_i$ is generally unknown, $\beta_i$ can be assumed to be proportional to $\alpha_i$. This is because large $\alpha_i$ in \eqref{eq:old_proposed} and $\beta_i$ in \eqref{eq:new_proposed} both lead to a dense graph. Therefore, we use $\beta_i\propto \sum_{j=0}^N[\mb{Z}]_{i,j}$.

\subsection{Computational Complexity}
\label{subsec:complexity}

We compare the computational complexity of the proposed method with those of representative existing methods\footnote{We omit the derivation due to page limitation.}.
The worst-case complexities are summarized in Table \ref{tab:complexity}.
The graph construction methods typically require 1) a preparation phase and 2) an optimization phase.
Below, we compare the complexity in both phases.

\paragraph*{Preparation}

The fixed $k$NNG construction has the computational complexity $O(N^2 D k_{\max})$.
NNK \cite{shekkizhar_graph_2019} requires the calculation of a fixed $k_{\max}$NNG in the initial step.
Graph learning in \eqref{eq:glasso} requires no calculation in the preparation phase.
The proposed method requires the calculation of $k_{\min}$NNG in the initial step.

\begin{figure}[t!]
\setlength{\abovecaptionskip}{5pt}
    \centering
    \includegraphics[width=1.0\linewidth]
          {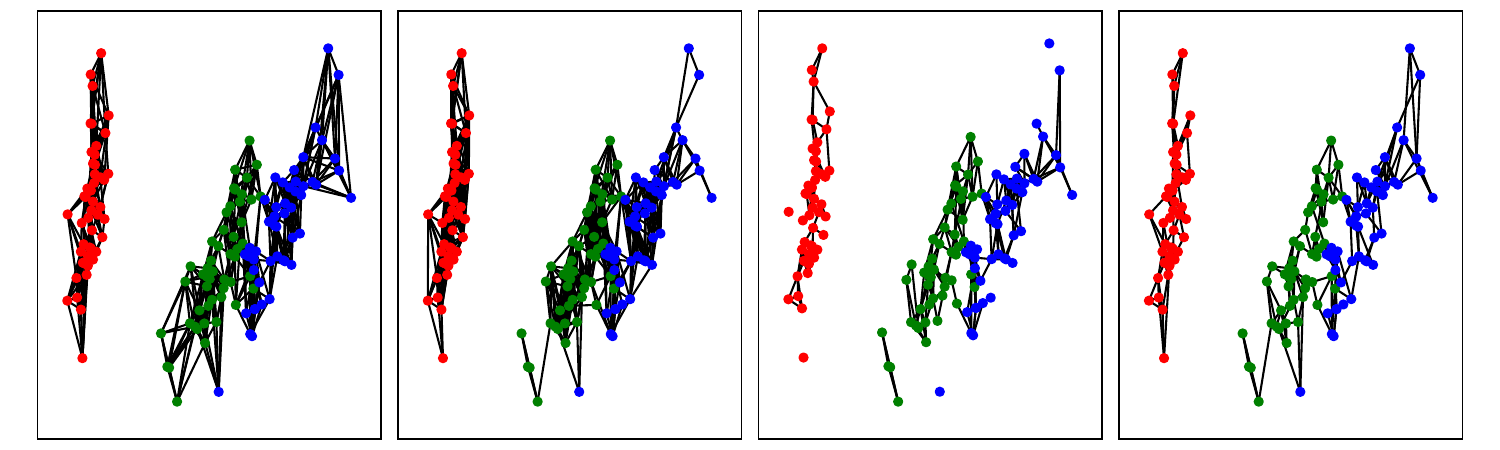}
   \caption{Graphs constructed from the iris dataset. From left to right: $k_{\max}$NNG, graph learning, NNK, proposed v$k$NNG with $(k_{\min},k_{\max})=(3,7)$.}
    \label{fig:iris_graph}
\end{figure}

\paragraph*{Optimization}

In the optimization step, graph learning requires $O(N^5+N^{3\omega}+N^{2\omega})$ complexity where $\omega$ is a constant for SDP.
Clearly, solving \eqref{eq:glasso} is computationally demanding because no efficient algorithm for solving \eqref{eq:glasso} has been developed so far\footnote{An efficient method for a similar problem to \eqref{eq:glasso} based on the block coordinate descent has been proposed in \cite{zeng_sparse_2022}.}.
NNK requires $O(Nk_{\max}^3)$ complexity for solving quadratic programming methods \cite{slawski_non-negative_2013}. 
The proposed method needs $O(N^2D)$ to calculate the distance matrix and $O(N^2)$ for sorting $\mb{Z}$. In the worst-case scenario, finding the optimal $k$ requires $O(Nk_{\max})$.
As a result, the proposed method requires $O(N^2D+N^2+Nk_{\max})$, which is comparable to the complexity of the classical $k$NN approach.

\section{Experiments}
In this section, we validate the effectiveness of the proposed method with real datasets.
We compare the experimental results with existing methods introduced in Sec.~\ref{sec:related_work}.

\subsection{Visualization of Estimated Graphs}

We use the popular \textit{Iris} dataset\footnote{https://archive.ics.uci.edu/dataset/53/iris.}.
First, we embed the data into the 2-D space spanned by the two principal vectors obtained from principal component analysis. We visualize the estimated graphs in Fig.~\ref{fig:iris_graph}. 
As expected, graph learning produces a dense graph where controlling sparsity is usually difficult.
In contrast, we can see that the proposed method has a sparse but reasonably connected graph.
The graphs obtained by the proposed method and NNK are comparable but their computational complexities are different (see Section \ref{subsec:complexity}).

\subsection{Point Cloud Denoising}
We perform point cloud denoising via a graph low-pass filter since it scales with the number of points compared to deep learning approaches and it helps to highlight the differences of graph estimation methods.
Since the graph learning approach from  \eqref{eq:glasso} is too complex to be used for datasets of the size of point clouds (see Table \ref{tab:complexity}), we do not show its denoising result here.
The point clouds are taken from Modelnet10\footnote{https://modelnet.cs.princeton.edu/}.
In the dataset, we only use the objects having greater than 1000 points and we downsample them to 1000 points. The attributes are the 3-D coordinates normalized to $[0,1]$. Noisy coordinates are corrupted by Gaussian noise conforming to either $\mathcal{N}(0, 0.05)$ or $\mathcal{N}(0, 0.1)$.
We perform denoising with the heat kernel having its graph frequency response $e^{-50 \lambda}$.

To conduct the experiment under fair conditions, we tune the parameters of all methods so that the average degrees are close to 10. 
For $k$NNGs (both conventional and proposed), we use the radial basis kernel with $\gamma = 30$ as the metric for determining edge weights. 

The average mean squared errors (MSEs) in decibels for 10 independent runs are summarized in Table \ref{tab:exp_result}.
For all cases, the proposed v$k$NNG outperforms those of other methods.
The proposed method tends to connect similar nodes as many as possible, while NNK avoids connecting similar nodes if they are very close (in the feature space)\cite{shekkizhar_graph_2019}.
This characteristic of the proposed method can help the denoising application and could improve the MSE.

\begin{table}[]
\setlength{\tabcolsep}{0.30em}
\centering
\caption{Average MSEs for point cloud denoising (dB). $\sigma$ is the standard deviation of noise. The bold types denote the best MSEs for each $\sigma$.}\label{tab:exp_result}
\begin{tabular}{c|c|c|c|c|c|c}
\hline
Method         & \multicolumn{2}{c|}{$k$NNG}      & \multicolumn{2}{c|}{NNK} & \multicolumn{2}{c}{Proposed}          \\\hline
$\sigma$          & 0.05              & 0.1      & 0.05       & 0.1        & 0.05              & 0.1               \\\hline\hline
\texttt{bathtub}        & -26.33         & -22.65 & -27.17  & -23.06  & \textbf{-28.04} & \textbf{-23.30} \\
\texttt{bed}            & -26.02          & -22.27 & -26.69   & -22.66    & \textbf{-27.61} & \textbf{-22.94} \\
\texttt{chair}          & -26.15           & -22.08 & -26.78    & -22.46   & \textbf{-27.74} & \textbf{-22.86}  \\
\texttt{desk}           & -27.07          & -22.95  & -28.02   & -23.24   & \textbf{-28.57} & \textbf{-23.48} \\
\texttt{dresser}        & -27.62          & -23.00 & -28.50   & -23.21   & \textbf{-28.99} & \textbf{-23.46}  \\
\texttt{monitor}        & -26.66 & -22.59 & -27.82   & -23.10    & \textbf{-28.60} & \textbf{-23.17} \\
\texttt{nightstand}     & -27.57           & -23.58  & -28.53   & -23.86   & \textbf{-28.83} & \textbf{-24.05}  \\
\texttt{sofa}           & -25.15          & -21.35 & -25.93    & -21.83   & \textbf{-26.94} & \textbf{-22.18} \\
\texttt{table}          & -26.80          & -22.86 & -27.46   & -23.18   & \textbf{-28.11}   & \textbf{-23.51} \\
\texttt{toilet}         & -23.42           & -20.40 & -24.13   & -20.92   & \textbf{-25.82} & \textbf{-21.42} \\\hline\hline
Ave. degree & 12.82          & 12.76 & 10.89   & 10.93   & 10.65          & 10.77         \\
\hline
\end{tabular}
\end{table}

\section{Conclusion}
We propose a v$k$NN graph construction method
based on the graph learning approach.
We formulate a node-by-node optimization problem for determining the best $k$s in v$k$NN graphs.
We reveal a clear relationship between the proposed method and graph learning methods and discuss the parameter selection in our approach.
In experiments on real datasets, we demonstrate that the proposed method outperforms existing methods.

\newpage

\end{document}